\documentclass[runningheads]{llncs}

\usepackage[caption=false]{subfig}
\usepackage[shortlabels]{enumitem}
\usepackage{multirow}
\usepackage{amsmath}
\usepackage{float}
\usepackage{array}
\usepackage{lipsum,graphicx}
\usepackage{times}
\usepackage{fancyhdr,graphicx,amsmath,amssymb}
\usepackage[ruled,vlined]{algorithm2e}
\usepackage{algorithmic}
\include{pythonlisting}
\usepackage{booktabs}
\usepackage{float}

\begin{document}
\emergencystretch 3em 
\title{Facial Recognition in Collaborative Learning Videos\thanks{This material is based upon work supported by the National Science Foundation under Grant No.1613637, No.1842220, and No.1949230.}}
\author{Phuong Tran \inst{1} \and Marios Pattichis \inst{1}
        \and Sylvia Celed\'on-Pattichis \inst{2} 
        \and Carlos L\'opezLeiva \inst{2}}

\authorrunning{P. Tran et al.}
\institute{
Dept. of Electrical and Computer Engineering\\
University of New Mexico, Albuquerque NM 87106, USA 
\and
Dept. of Language, Literacy, and Sociocultural Studies\\
University of New Mexico, United States.
}

\maketitle              
\begin{abstract}
Face recognition in collaborative learning videos presents many challenges.
In collaborative learning videos, students sit around a typical table at different positions to the recording camera, come and go, move around,
   get partially or fully occluded. 
Furthermore, the videos tend to be very long, requiring
   the development of fast and accurate methods.
   
We develop a dynamic system of recognizing participants in collaborative learning systems.
We address occlusion and recognition failures by using
   past information about the face detection history.
We address the need for detecting faces from different poses and the need for speed
   by associating each participant with a collection of prototype faces computed
   through sampling or K-means clustering.

Our results show that the proposed system is proven to be very fast and accurate.
We also compare our system against a baseline system that uses InsightFace \cite{insightface} 
   and the original training video segments.
We achieved an average accuracy of 86.2\% compared
    to 70.8\% for the baseline system.
On average, our recognition rate was 28.1 times faster than the baseline system.

\end{abstract}

\begin{figure}[!t]
	\centering
	\includegraphics[width=1\textwidth]{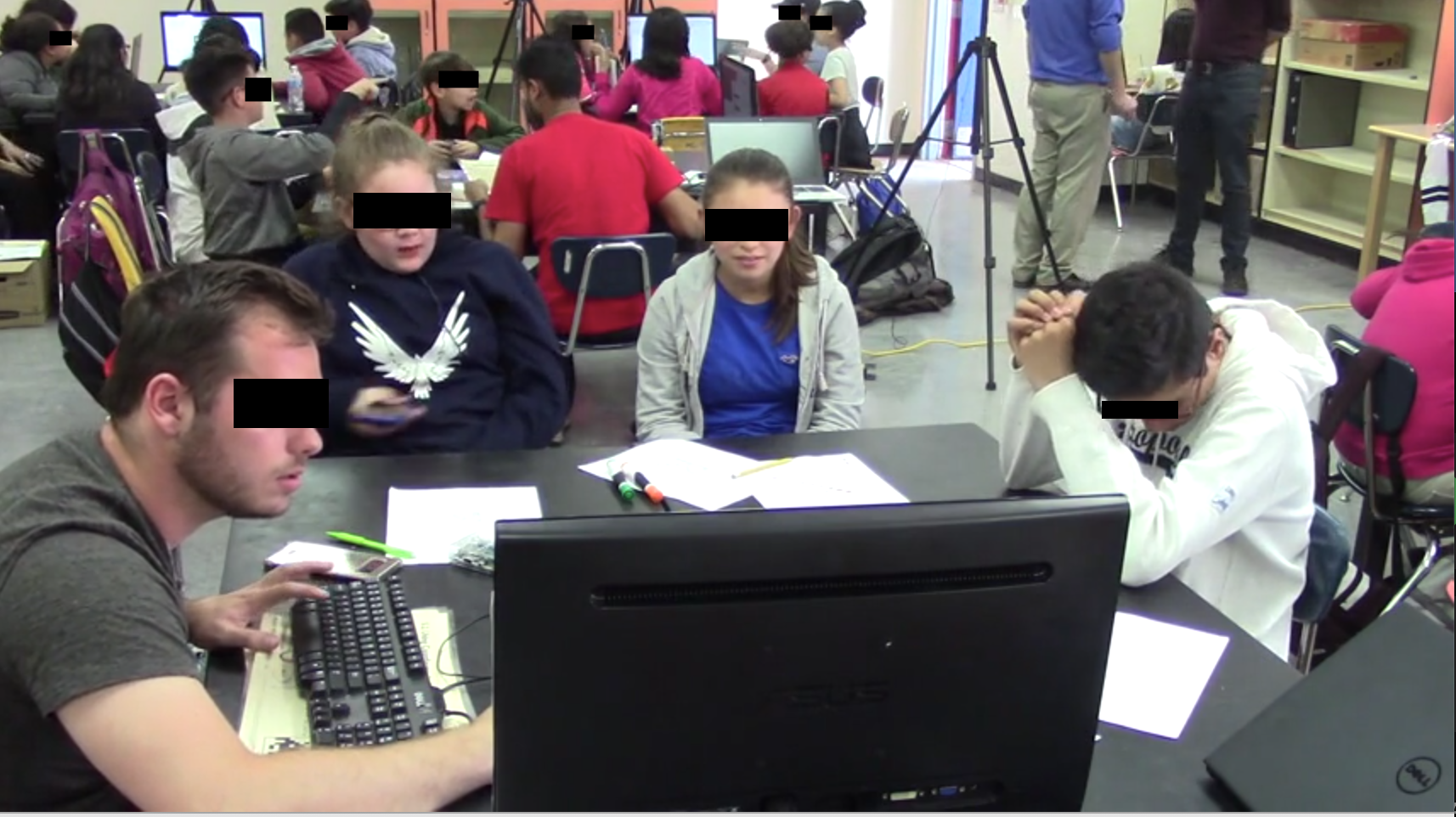}
	\caption{
	Example of a collaborative learning group.
	We seek to recognize the students who are
	closer to the camera while ignoring students
	from all other groups. 
	} 
	\label{fig:eg}
\end{figure}
\section{Introduction}
We study the problem of face recognition in collaborative learning
   environments.
Our goal is to develop fast and accurate methods that
   can be used to quantify student participation
   as measured by their presence in their learning groups.
   
Fig. \ref{fig:eg} represents an example of 
     a collaborative learning group. 
A collaborative learning group 
     is represented by the group of students
     closest to the camera.
Background groups are not considered part of
     the collaborative group that we are analyzing.
There is a possibility that students or
     facilitators move between groups.
Thus, we need to recognize the current
      members of our group
      from a larger group of students.

A fundamental challenge of our dataset is that 
      face recognition needs to work
      at a large variety of poses.
As long as a sufficiently small part
      of the face is visible,
      we need to identify the student.
As an example, in Fig. \ref{fig:eg}, the student
      on the front right with a white hoodie
      has his face mostly blocked. 
Furthermore, students may disappear or reappear
      because the camera moves, or the students take
      a break, or because they have to leave the group.
Hence, there are significant video properties
      to our problem that are not present
      in standard face recognition systems.

As part of a collaborative project between engineering 
   and education, our goal is to assist educational researchers in analyzing how students who join the program learn and facilitate the learning of other students.
The problem of identifying who is crucial 
   for assessing student participation in the project.
Furthermore, the developed methods
   need to be fast.
Eventually, we will need to apply our methods to 
   approximately one thousand hours
   of videos that we need to analyze.

The standard datasets for face recognition
   use a database of front-facing images.
The Labeled Faces in the Wild (LFW) dataset \cite{LFWTech} 
   contains more than 13,000 face images with various poses, ages, and expressions. 
The Youtube Face (YTF) dataset \cite{LFWTechUpdate}
   contains around 3,500 videos with an average range of 181 frames/video from 1,600 different people.
The InsightFace system \cite{Deng} developed the use of 
   Additive Angular Margin Loss for Deep Face Recognition 
   (ArcFace) on a large-scale image database with trillions of pairs 
   and a large-scale video dataset, and tested on multiple datasets with different loss function models (ArcFace, Softmax, CosFace,..). 
InsightFace gave the best accuracies on 
    LFW and YTF with 99.83\% 98.02\%.
We adopt InsightFace as our baseline face 
    recognition system because of 
    their state-of-the-art performance.

We also provide a summary of video analysis methods
    that were developed specifically by our group for
    analyzing collaborative learning videos.

In \cite{Abby2018},
    the authors introduced methods for detecting
    writing, typing, and talking activities using motion vectors and deep learning.
In \cite{Shi2018}, 
   the authors developed methods
   to detect where participants
   were looking at. 
In \cite{Luis2019}, 
   the authors demonstrated that FM images with low-complexity neural networks 
   can provide face detection results that
   can only be achieved with much more complex deep learning systems.
   
We provide a summary of the contributions of the
       current paper.
First, we introduce the use of
       a collection of face prototypes
       for recognizing faces from different angles.
Second, we apply multi-objective optimization to
       study the inter-dependency between recognition rate,
       the number of face prototypes, and recognition accuracy.
Along the Pareto front of optimal combinations,
       we select an optimal number of face prototypes 
       that provides for a fast approach
       without sacrificing recognition accuracy.
Third, we use the past recognition history to deal with
       occlusions and, hence, support
       consistent recognition throughout the video.
Compared to InsightFace \cite{insightface}, the proposed
       system provides for significantly faster
       recognition rates and higher accuracy.

We summarize the rest of the paper into three additional sections.
In section \ref{sec:Meth}, we provide a summary of our methodology.
We then provide results in section \ref{sec:Res} and concluding
   remarks in section \ref{sec:Con}.

\section{Methods}\label{sec:Meth}
\begin{figure}[!t]
\centering
 \includegraphics[width=.7\textwidth]{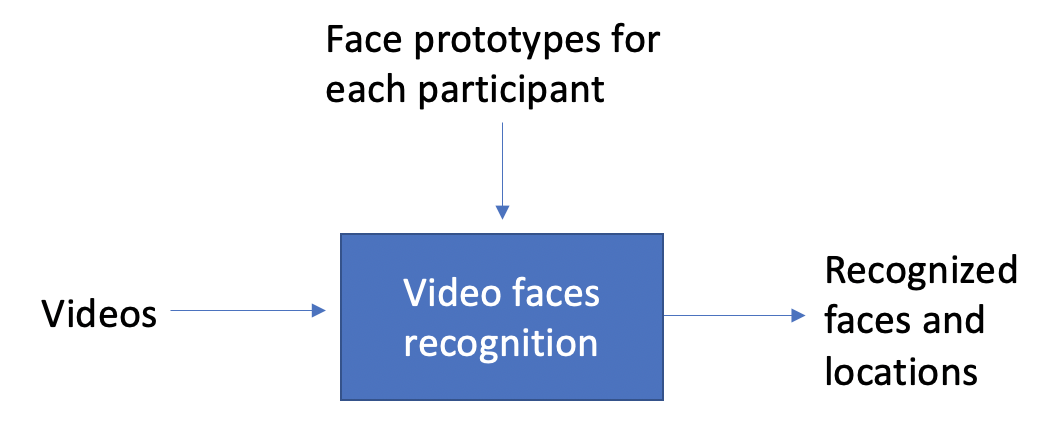}
 \label{fig:topDown}

\begin{algorithm}[H]
\SetAlgoLined
\textbf{Input:}\\
$\quad$ video clips associated with each participant.\newline
\textbf{Output:}\\
$\quad$ facePrototypes associated with each participant.\newline
\textbf{for} \text{each participant} \\
  \quad$\textbf{Apply} \text{ K-means clustering}$ \\
  \quad$\textbf{Select} \text{ cluster means}  $ \\
  \quad$\textbf{Find} \text{ nearest images from cluster centroids}$\\
  \quad$\textbf{Align} \text{ faces to 112x112}$\\
\textbf{end}
\label{k-means}
\caption{Computing Face prototypes using K-means}
\end{algorithm}
\label{fig:systemDiag}
 \caption{
 Block diagram for recognizing faces from videos.
 Each face is associated with a collection of face prototypes.
The method computed face prototypes using video sampling
      or K-means clustering.
 K-means clustering is given here.}
\end{figure}

We present a block diagram of the overall system in
   Fig. \ref{fig:systemDiag}. 
Our video recognition system requires
   a set of face prototypes associated with
   each participant.
The video face recognition algorithm
   detects the faces
   in the input video and computes
   minimum distances to the face prototypes
   to identify each participant.
To handle occlusion issues,
   the system uses past detection history.

\begin{algorithm}[!htp]
    \SetAlgoLined
    \textbf{Input:}\\
    $\quad$ {\tt video}: unlabeled video\\
    $\quad$ {\tt facePrototypes}: list of images associated with
            each student pseudonym\\
    \textbf{Output:}\\
    $\quad$ {\tt vidResult}: store unique student identifiers and face locations along with face's landmarks per frame\\
    \textbf{Local Variables:}\\
    $\quad$  {\tt ActiveSet} and {\tt InactiveSet} store unique student
    identifiers, face locations, {\tt totalAppearances}, \\
    $\qquad$ {\tt totalFramesProcessed}, {\tt continuousAppearances} \& distance.\\[0.05in]
    
    \textbf{while} \text{frame {\tt f} in initial part of video} $\rhd\text{Detect and Recognize all faces initial duration}$ \\
      $\quad$\textbf{Detect} faces in {\tt f}\\
      $\quad$\textbf{Recognize} faces in {\tt f} using minimum distance to {\tt facePrototypes}\\
      $\quad$\textbf{Update} {\tt vidResult}\\[.1in]
    \textbf{{\tt ActiveSet} = []; {\tt InactiveSet} = [];} $ \rhd\text{Initialization}$ \\
     \textbf{while} {\tt face} in all recognized faces \\
     $\quad${\bf if} {\tt totalAppearances}({\tt face}) / {\tt totalFramesProcessed}({\tt face}) $>=$ 50\%\\
     $\qquad${\bf Add} {\tt face} to ActiveSet \\
     $\quad${\bf else}:\\
     $\qquad${\bf Add} {\tt face} to InactiveSet\\[0.05in]
      \textbf{for} \text{frame {\tt f} in rest of {\tt video}}\\
      $\quad$\textbf{Detect} faces in {\tt f}\\
      $\quad$\textbf{if} \text{minor movement in detected face} \textbf{then}  $\quad\rhd\text{Reuse face}$\\
      $\qquad$\textbf{Reuse} face from  {\tt ActiveSet} \\
      $\qquad${\bf Update} {\tt ActiveSet}, {\tt vidResult} with detected faces  \\
      $\quad$\textbf{else if} \text{detected face found in {\tt InactiveSet}} \textbf{then} $\quad\rhd\text{Update face}$ \\
      $\qquad${\bf Update} {\tt InactiveSet} with detected face \\
      $\qquad${\bf if} {\tt totalAppearances}({\tt face}) / {\tt totalFramesProcessed}({\tt face}) $>=$ 50\%\\
      $\quad\qquad${\bf Move} face to {\tt ActiveSet} \\
      $\quad\qquad${\bf Remove} face from {\tt InactiveSet} \\
      $\quad\qquad${\bf Update} {\tt ActiveSet}, {\tt vidResult} with detected faces \\ 
      $\quad$\textbf{else} $\quad\rhd\text{Possible new face}$\\
      $\qquad$\textbf{Recognize} face in f using minimum distance to {\tt facePrototypes}\\
      $\qquad${\bf Update} {\tt InactiveSet} with detected faces \\ [0.05in]
      $\quad$\textbf{for} \text{face in  {\tt ActiveSet}}\\
      $\qquad$\textbf{if} \text{ face not found in all detected faces} \textbf{ then} \\
      $\quad\qquad$\textbf{if}  {\tt continuousAppearances} $>=$ {\tt minAppearances} \textbf{then}  $\quad\qquad\rhd{\text{Occluded face}}$ \\
      $\qquad$ $\qquad$\textbf{Add} \text{face to  {\tt ActiveSet}, {\tt vidResult}}\\
      $\quad$ $\qquad$\textbf{if}  {\tt continuousAppearances} $<$ {\tt minAppearances} \textbf{then} $\quad\qquad\rhd\text{Disappearing face}$ \\
          $\qquad$ $\qquad$\textbf{Remove} face from  {\tt ActiveSet} \\
      $\qquad$ $\qquad$\textbf{Add} face to  {\tt InactiveSet}\\[0.05in]
      $\quad$ $\textbf{for} \text{ all labels found in {\tt {f}}}$ $\quad \rhd\text{Consistent assignment check}$\\
      $\qquad$ $\textbf{if} \text{ same label exists}$\\
      $\qquad \quad \textbf{Set} \text{ label with larger distance to Unknown}$\\
    	\caption{Video Faces Recognition} 
    	\label{videoProcessing}
\end{algorithm}

\begin{figure}[!b]
	\centering
	\includegraphics[width=1\textwidth]{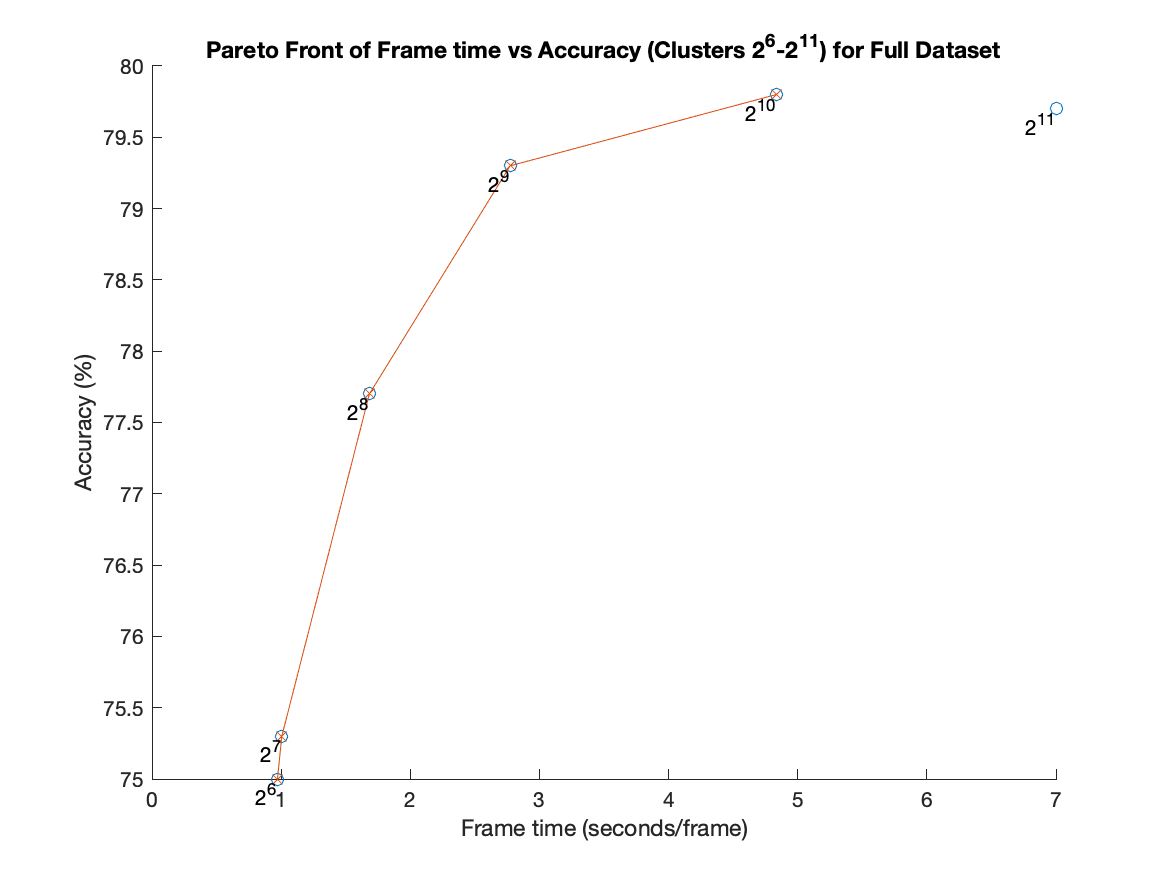}
	\caption{Pareto Front on the K-mean Clustering Results} 
	\label{fig:ParetoFront}
\end{figure}

\subsection{Computation of Face Prototypes}\label{sec:clusters}
We use two different methods to compute 
   the face prototypes.
First, we consider the use of K-means clustering
   to select a limited number of face prototypes 
   from the training videos.
Second, we consider a baseline approach where we
   use sparse sampling of the training videos to define
   the face prototypes.
To achieve sparsity, we use one sample image per second
   of video.
For our second approach, we used
   a video from Group D from
   urban cohort 2 level 1 (C2L1, Video 4) 
   and Group E from C3L1 (Video 5), whereas the first approach tested on Group C from urban C1L1 (Video 1).
Each sampled face is aligned and resized
   to $112\times 112$.

We summarize the K-means approach in 
   Fig. \ref{fig:systemDiag}.
We use K-means to cluster similar frames
   that appear when a student does not move very much.
Hence, we expect that the centroids that result
   from the K-means algorithms would represent
   a small number of diverse face poses for each participant.
To avoid unrealistic centroid images, we compute our face prototypes using the images closest to our cluster centroids.
After finding a prototype image that
   is closest to the mean,
   we align and resize each prototype image
   to $112\times 112$.

We use multi-objective optimization to determine
   the minimum number of face prototypes without sacrificing classification accuracy.
We present our results for a representative
   video from group C, level 1, in Fig. \ref{fig:ParetoFront}.
For the optimization,
   we use a log-based search by
   varying the number of face prototypes
   from 2$^0$ to 2$^{11}$. 
For K-means clustering,
   we see that the accuracy
   peaks at 79.8\% 
   for 1024 face prototypes, with a recognition
   rate of 4.8 seconds per frame.  
   In this example, we only used one video to validate our approach. 
   We expand the validation set to more videos as described in \cite{thesis}. This process run on a personal 
   Macbook Pro that ran Mac-OS with 2.3GHz, 4-core, Intel i5 processors.
   
\subsection{Video Faces Recognizer}\label{sec:videorec}
We present the algorithm for video face recognition
   in Algorithm \ref{videoProcessing}. 
The input is an unlabeled video and the {\tt facePrototypes} that  
   provides a list of images associated with each 
   participant. 
The detected faces for
   each video frame are returned
   in {\tt vidResult}.
To address occlusion, the algorithm
   maintains the face recognition history
   in {\tt ActiveSet} and {\tt InactiveSet}.
   
First, for the first two seconds of 
   the videos, we detect all participants
   in each video frame using
   MTCNN \cite{MTCNN}.
For each detected face, MTCNN computes
   five landmark points:
   two for the eyes, one for the nose, and two for the 
   mouth corners.
The face detector uses a minimum area requirement
   to reject faces that belong to another group. 
Thus, we reject smaller face detections that are part of another group or other false positives because they appear smaller in the video.
Second, we classify each detected face by
   selecting the participant that gives
   minimum distance to their associated prototypes
   stored in {\tt facePrototypes}. 

We use the initial face recognition results to
   initialize {\tt ActiveSet} and {\tt InactiveSet}.
The faces that have been
   recognized in more than half of the frames
   are stored in the {\tt ActiveSet}.
The rest of them are stored in the 
   {\tt InactiveSet}.
For each face detection, we use a dictionary to store:
   a pseudonym, location information,
   \mbox{\tt totalAppearances} that stores the total number of frames where this face was detected, and {\tt totalFramesProcessed} which represents the total number of frames processed since this face appeared for the first time.
Hence, we use the {\tt ActiveSet} to hold
   faces that appear consistently whereas {\tt InactiveSet} contains faces that are still in doubt.

When a recognized face enters the 
   {\tt ActiveSet}, it gets a maximum value
   of 10 for its corresponding
   {\tt continuousAppearances}. 
When a previously recognized face is missing,
  {\tt continuousAppearances} gets decremented
  by 1.
When a face re-appears,
  {\tt continuousAppearances} is incremented
  until it reaches 10 again.
We also set {\tt minAppearances} to 5 as the minimum
  requirement on the number of prior continuous appearances for
  addressing occlusion issues.
Thus, for each face in the
  {\tt ActiveSet} that is not being detected in any frame,
  if  {\tt continuousAppearances} $\geq$ {\tt minAppearances},
  we declare the face as occluded, we mark it as present, and update {\tt vidResult}.
Else, if {\tt continuousAppearances} $<$ {\tt minAppearances},
  we declare the face as disappearing, and move it to the 
  {\tt InactiveSet}. 

We thus process the rest of the 
   {\tt video}
   based on the following four cases:
\begin{enumerate}[topsep=0pt]
 \item[(i)] If a newly detected face corresponds to a minor movement of a prior detection,
   we keep it in the {\tt ActiveSet}. This approach leads to a significant speedup in our face recognition speed.
 \item[(ii)] If a newly detected face is in the {\tt InactiveSet},
   we update the {\tt InactiveSet} with the new detection, and
   look at the ratio of {\tt totalAppearances}/{\tt totalFr- amesProcessed}
   to determine if it needs to move to the {\tt ActiveSet}. Otherwise, the face stays in {\tt InactiveSet}.
 \item[(iii)] If a newly detected face does not belong to either set,
  then recognize it and move it to the {\tt ActiveSet}.
 \item[(iv)] If a face that belongs to the {\tt ActiveSet} no longer appears,
  we consider the case for occlusion and that the participant has left the frame. As described earlier, we check {\tt continuousAppearances} to determine whether to declare the face
  occluded or not.
\end{enumerate}   
  
Lastly, we do not allow the assignment of the same label to two different faces in the same frame.
In this case, the face that gives the minimum distance is declared the recognized face while the other(s) are declared Unknown.

\begin{table}[!t]
	\caption{\label{AccRes}
	Accuracy for Facial Recognition. 
	Each video represents a different group session segment.}
		\begin{center}
		\begin{tabular}{p{0.28\textwidth}p{0.15\textwidth}p{0.2\textwidth}p{0.15\textwidth}p{0.15\textwidth}}
			\toprule
			\textbf{Video}  & \textbf{Duration}                  & \textbf{Person Label}           & \textbf{Ours} & \textbf{Insightface} \\ \toprule
			\multirow{4}{*}{\begin{tabular}[c]{@{}l@{}}\textbf{1}\\ (Face prototypes using\\ K-means)\end{tabular}} & \multirow{4}{*}{10 seconds}&\textit{\textbf{Antone39W}}& \textbf{36.5\%}    & \textbf{36.5\%}      \\
			 & & \textit{\textbf{Jaime41W}}   &   \textbf{86.7\%}   & 84.2\%       \\
			& & \textit{\textbf{Larry40W}}  &    \textbf{99.3\%}                          & 98.3\%         \\
			& & \textit{\textbf{Ernesto38W}}   &    \textbf{96.5\%}                          & 95.3\%
			\\ \cmidrule{3-5}
			& & \textbf{Average}   & \textbf{79.8\%}                          & 78.6\%      \\ \midrule
			
			\multirow{4}{*}{\begin{tabular}[c]{@{}l@{}}\textbf{2}\\ (Face prototypes using \\Sampling)\end{tabular}}  & \multirow{4}{*}{10 seconds}& \textit{\textbf{Chaitanya}}          &\textbf{95.3\%}    & 80.3\%       \\
			&& \textit{\textbf{Kenneth1P}}     & \textbf{91\%}                          &83.1\%                    \\
			&& \textit{\textbf{Jesus69P}}   & \textbf{100\%}                         &\textbf{100\%}                         \\
			&& \textit{\textbf{Javier67P}}    & \textbf{100\%}                          &69.1\%                    
			\\ \cmidrule{3-5}
			&& \textbf{Average}   & \textbf{96.5\%}                          & 83.1\%           \\ \midrule
			
			\multirow{4}{*}{\begin{tabular}[c]{@{}l@{}}\textbf{3}\\ (Face prototypes using \\Sampling)\end{tabular}}  & \multirow{4}{*}{60 seconds} & \textit{\textbf{Chaitanya}}          &\textbf{80.0\%}    & 56.1\%       \\
			&& \textit{\textbf{Kenneth1P}}     & \textbf{98.3\%}                          &61.5\%                    \\
			&& \textit{\textbf{Jesus69P}}   & \textbf{99.3\%}                         &\textbf{96.3\%}                         \\
			&& \textit{\textbf{Javier67P}}    & \textbf{80.6\%}                          &39.0\%                    
			\\ \cmidrule{3-5}
			&& \textbf{Average}   & \textbf{89.5\%}                          & 63.2\%           \\ \midrule
			
			\multirow{4}{*}{\begin{tabular}[c]{@{}l@{}}\textbf{5}\\ (Face prototypes using \\Sampling)\end{tabular}}  & \multirow{4}{*}{10 seconds} & \textit{\textbf{Melly77W}}          &\textbf{96.0\%}    & 59.7\%       \\
			&& \textit{\textbf{Marisol112W}}     & \textbf{84.0\%}                          &60.5\%                    \\
			&& \textit{\textbf{Cristie123W}}   & 8.67\%                       &\textbf{27.3\%}                       \\
			&& \textit{\textbf{Phuong}}    & \textbf{77.4\%}                          &21.4\%                    
			\\ \cmidrule{3-5}
			&& \textbf{Average}   & \textbf{66.5\%}                          & 42.2\%            \\ \midrule
			
			\multirow{3}{*}{\begin{tabular}[c]{@{}l@{}}\textbf{5}\\ (Face prototypes using \\Sampling)\end{tabular}}  & \multirow{3}{*}{60 seconds} & \textit{\textbf{Alvaro70P}}          &\textbf{96.4\%}    & 60.8\%       \\
			&& \textit{\textbf{Donna112P}}     & \textbf{100\%}                          &99.8\%                    \\
			&& \textit{\textbf{Sophia111P}}   & 99.5\%                       &\textbf{99.9\%}                       \\
		     \cmidrule{3-5}
			&& \textbf{Average}   & \textbf{98.6\%}                          & 86.8\%            \\ \midrule
			
			&&\textbf{Overall Average}                             & \textbf{86.2\%}         & 70.8\%    \\
			
			\bottomrule
		\end{tabular}
	\end{center}
\end{table}

\section{Results}\label{sec:Res}
We sampled twenty-four participants from our video 
   dataset (11 boys and 13 girls). 
For training, we used 80\% of the data for fitting and 20\% 
   for validation using different video sessions.
From our video sessions, we randomly select
     short video clips of 5 to 10 seconds for training 
     our video face recognizer.
Overall, we use more than 200,000 video 
    frames from 21 different video sessions.
We have an average of about 10,000 image examples per participant.
Furthermore, as the program lasted three years, the testing dataset used later on videos (e.g., later cohorts and different levels).
For the testing dataset, we used seven video clips
   with a duration of 10 to 60 seconds.
However, we do assume that we have trained for all of the 
   students within the collaborative group.
For reporting execution times, we use
   a Linux system with an
   Intel(R) Xeon(R) Gold 5118 CPU running @ 2.30GHz
   with 16GB Memory and Nvidia Quadro RTX 5000 GPU with
   3072 Cuda cores.

We present face recognition accuracy results in
   Table \ref{AccRes} using said system.
For InsightFace,
   we use all of the training video frames 
   as face prototypes.
From the results,
   it is clear that the proposed method is far
   more accurate than the baseline method.
The difference in accuracy ranged from as low as 
   $2\%$ to as large as $25\%$. 
Out of 19 participants in these five videos, our 
   method achieved higher or same accuracy in 17 cases.
Overall, our method achieved $86.2\%$ 
   compared to $70.8\%$ for the baseline method. 
   
\begin{figure*}[!t]
	\centering
	(a)~\includegraphics[width=0.25\textwidth]{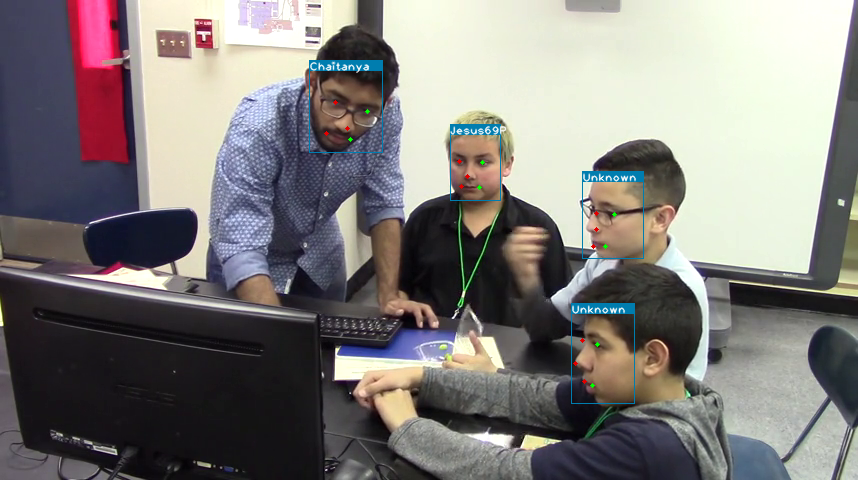}~
	(b)~\includegraphics[width=0.25\textwidth]{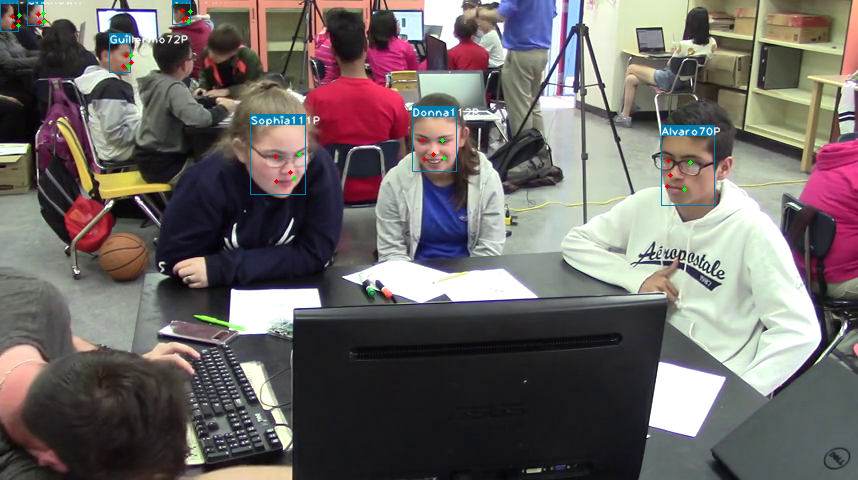}~
	(c)~\includegraphics[width=0.25\textwidth]{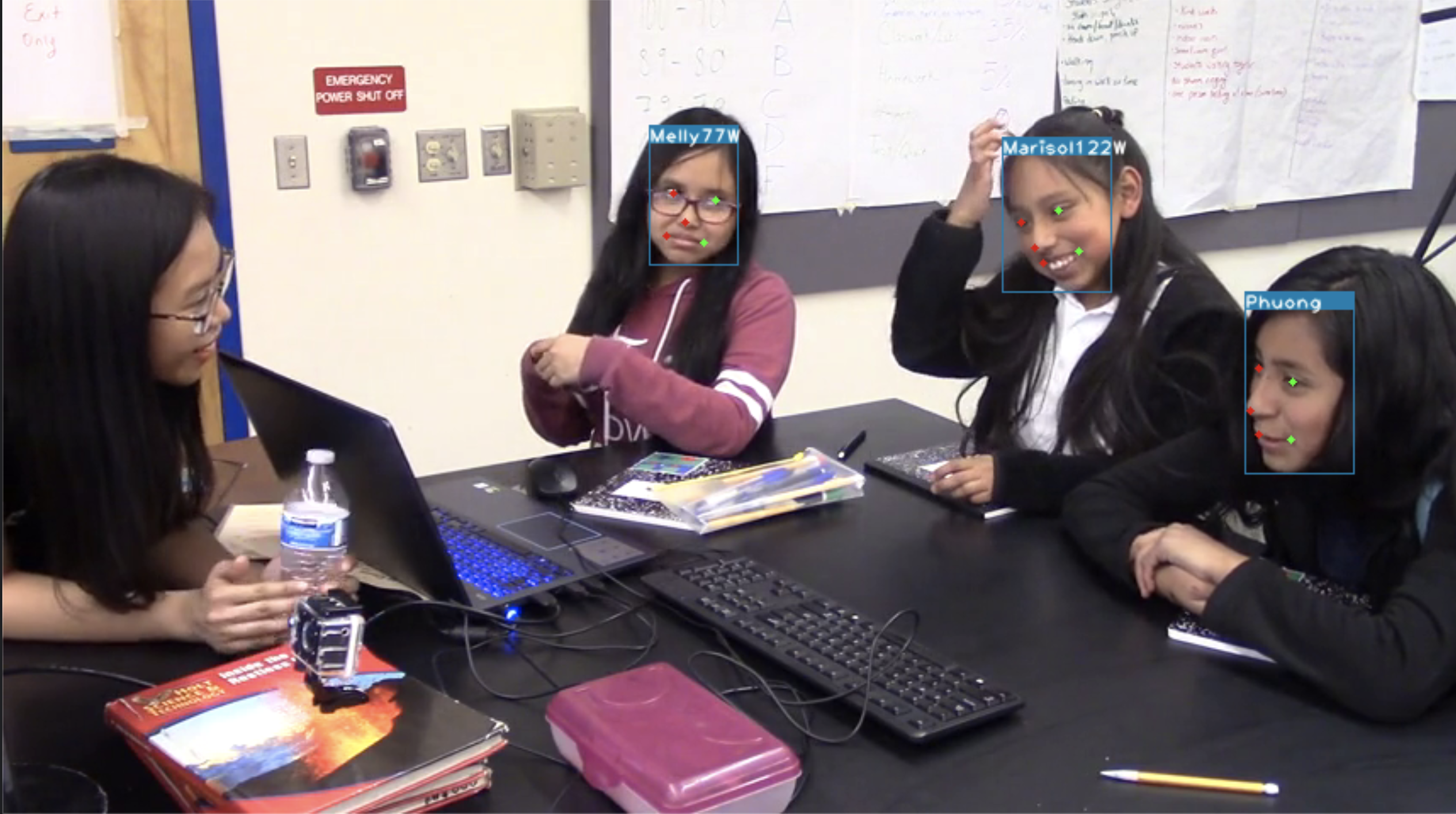}~\\[0.05 true in]
	(d)~\includegraphics[width=0.25\textwidth]{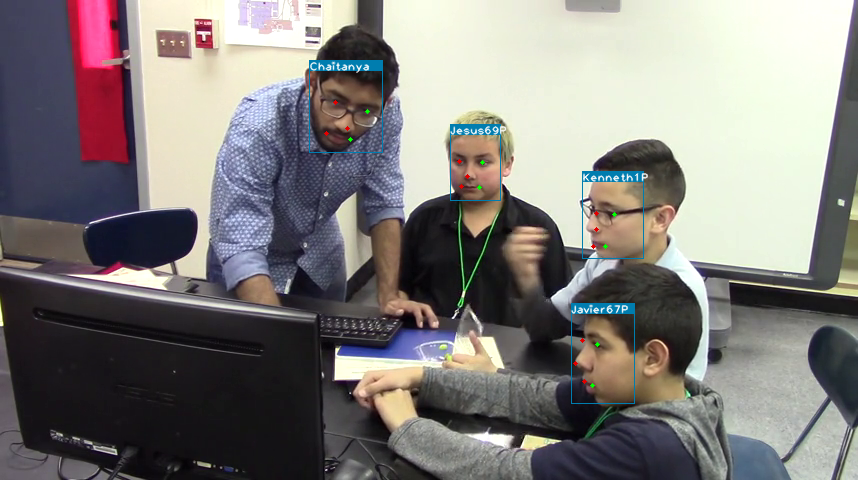}~
	(e)~\includegraphics[width=0.25\textwidth]{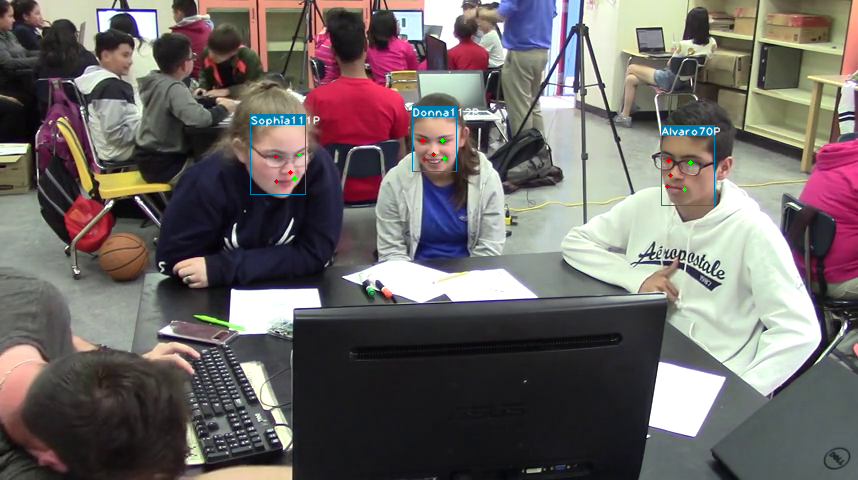}~
	(f)~\includegraphics[width=0.25\textwidth]{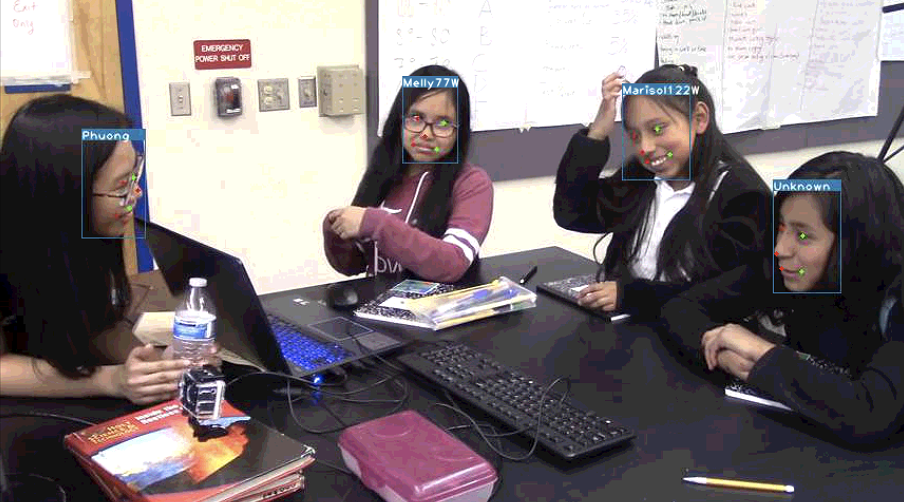}~\\[0.05 true in]
    (g)~\includegraphics[width=0.25\textwidth]{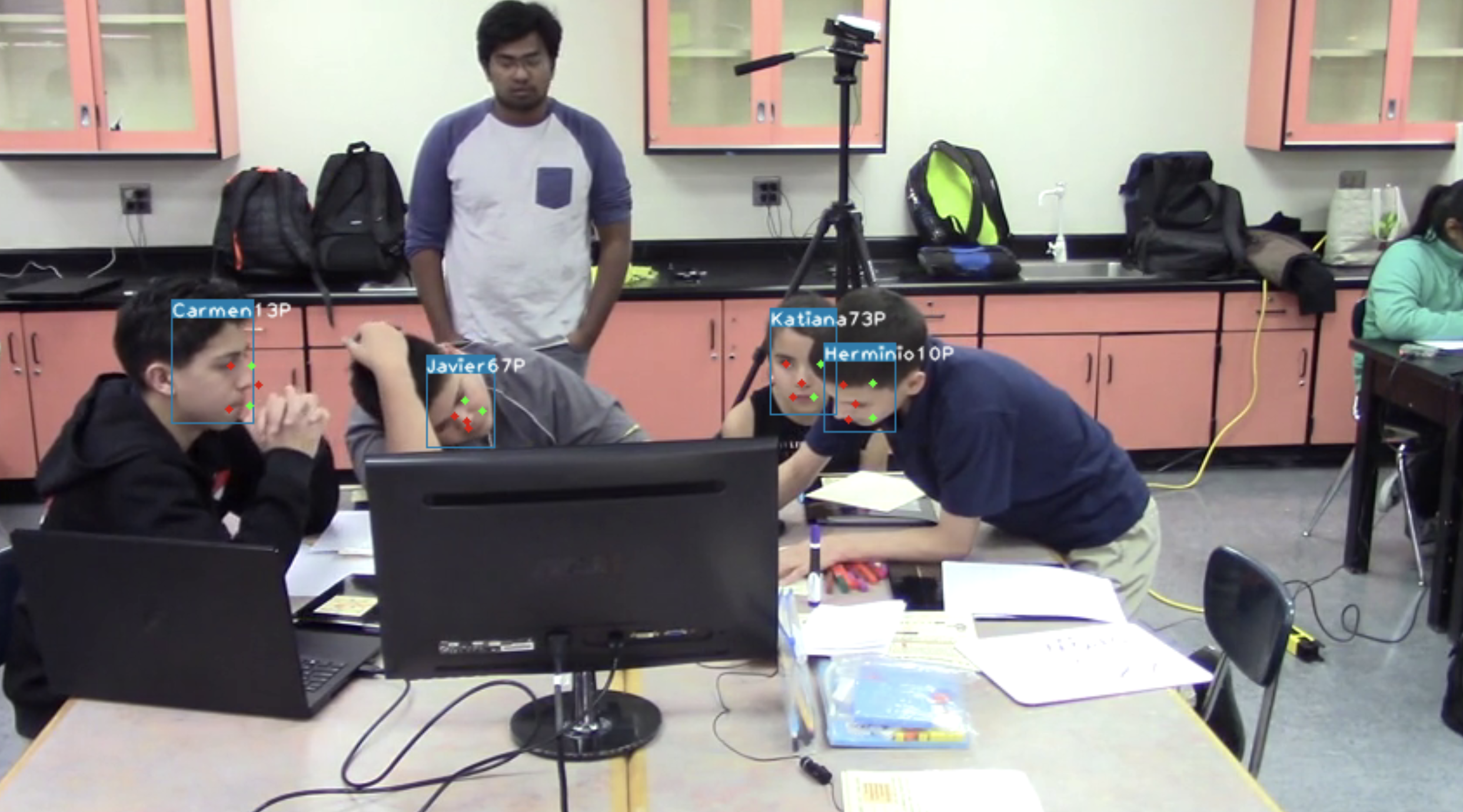}~
	(h)~\includegraphics[width=0.25\textwidth]{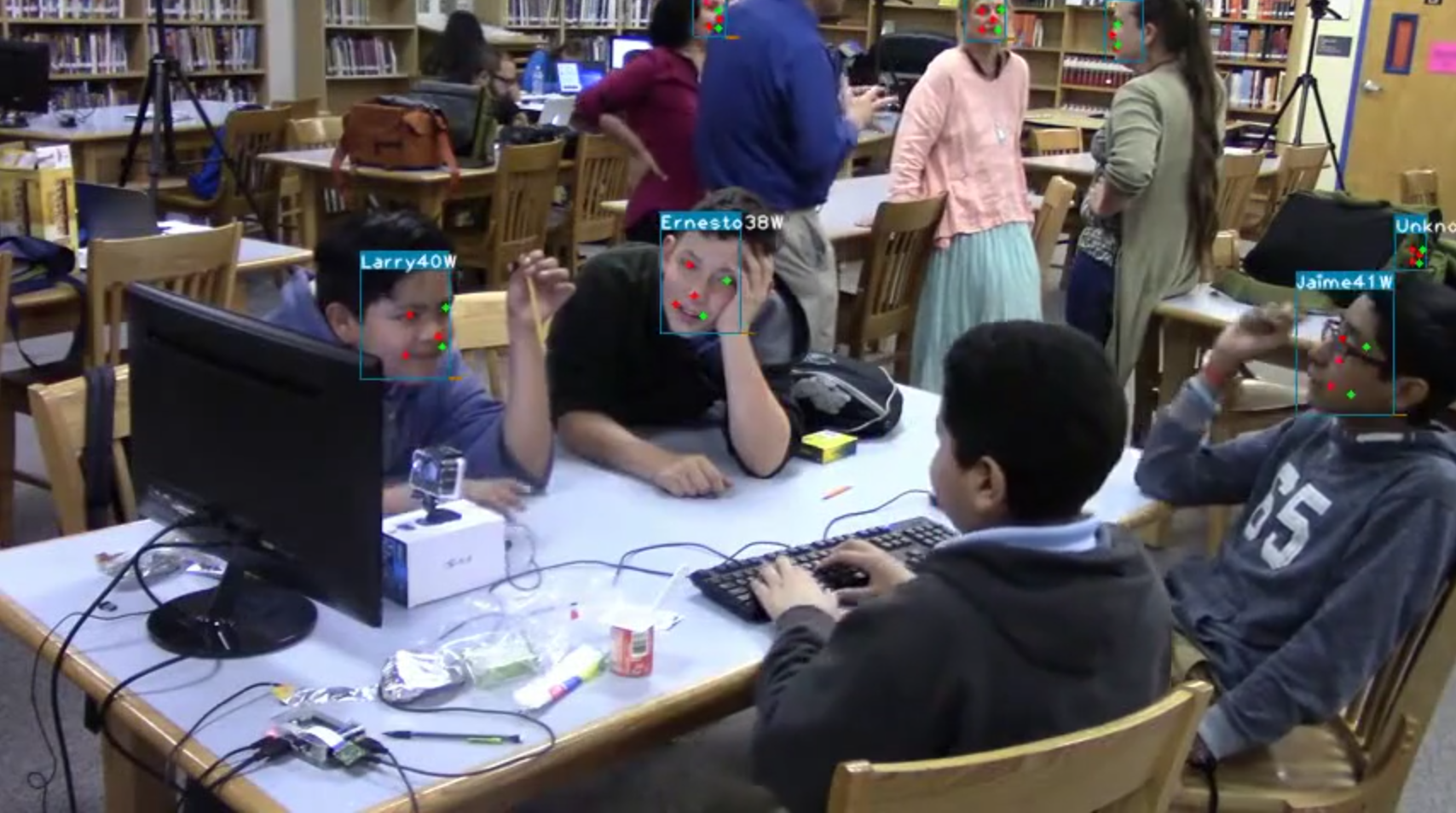}~
	(i)~\includegraphics[width=0.25\textwidth]{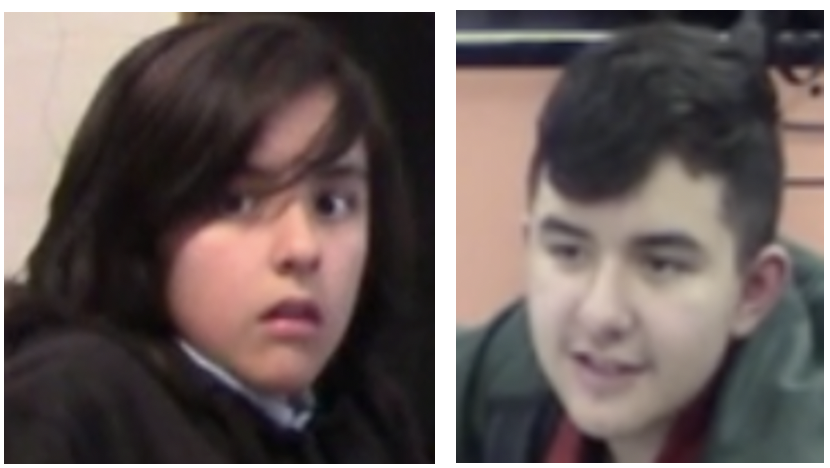}~\\[0.05 true in]
	\caption{
	Video face recognition results for three collaborative groups.
	The first row shows results from the use of InsightFace (baseline).
	The second row shows our results using the sampling method.
	In (g), we show successful detections despite 
	    occlusions.
	Results from the use of K-means clustering are shown in (h).
	Then, we show dramatic changes in appearance in (i).
	} 
	\label{fig:Interaction_sample}
\end{figure*}

We present face recognition examples in
   Fig. \ref{fig:Interaction_sample}.
Figs. \ref{fig:Interaction_sample}(a) 
   and \ref{fig:Interaction_sample}(b)
   show results from the same video frame of 
   Video 2. 
The baseline method recognized Javier67P (lower right in (a))
   and Kenneth1P (white shirt with glasses in (a)) 
   as Unknown,
   whereas our approach correctly identified all four 
   participants. 
A second example for video 5 is
   presented in Figs. \ref{fig:Interaction_sample}(b) and (e).
The baseline method identified all three people correctly.
However, the baseline method also detected and incorrectly claimed recognition of background participants that we did not train.
Our proposed method used projection and small-area elimination to reject this false-positive recognition.
A third example for video 4 is shown in
   Figs. \ref{fig:Interaction_sample}(c) and (f).
The baseline method 
   only succeeded in recognizing Melly77W (pink sweater) 
   and wrongly recognized Cristie123W (lower right) as Phuong, who is actually on the far left wearing
   glasses. 
Our method used history information to 
   address the partial occlusion issue
   and correctly recognized Phuong
   who is in the far left of 
   Fig. \ref{fig:Interaction_sample}(f).
Furthermore, our method rejected the wrong
   assignment of Phuong because
   it does not allow the assignment
   of the same identifier to two
   different faces.
Instead, the wrong assignment was
   re-assigned to Unknown.
A fourth example of our method is shown in 
   Fig. \ref{fig:Interaction_sample}(g).
In Fig. \ref{fig:Interaction_sample}(g),
   we can see that 
   our method works in occlusion cases. 
   Herminio10P (dark blue polo, right) 
   and Guillermo72P (blue T-shirt) were
   correctly recognized even though their faces
   were partial.
We also present challenges in 
   Figs. \ref{fig:Interaction_sample}(h) and (i).
In Fig. \ref{fig:Interaction_sample}(h),
  Antone39W did not get recognized because he had
  his back facing the camera.
In Fig. \ref{fig:Interaction_sample}(i),
  Kirk28P was not recognized due to significant
  changes in appearance through time.
   
We present speed performance comparisons
   in  Table \ref{TimeRes}.
The baseline method required 9.9 to 18.2 seconds/frame
  whereas our proposed method required 
  0.3 to 2.8 seconds/frame. 
On average, the proposed method
  was 28.1$\times$ faster.
Our speedups can be attributed to our use of
  a reduced number of face prototypes and the fact 
  that we do not rerun the minimum distance classifier 
  if there is little movement in the detected faces.
For example, for 3 and 4, 
  InsightFace took a very long time (more than ten seconds/frame) 
  because it compared each participant against (almost) ten thousand images.
For 5, in addition to comparisons to about ten thousand images for the main group,
  InsightFace also had to compare against faces from the background groups.
In comparison, our approach rejected the need to recognize background groups
  by applying a minimum face size constraint.
\begin{table}[!t]
	\caption{\label{TimeRes}
    	Recognition time for facial recognition. Each video represents a different group session segment.}
	\begin{center}
		\begin{tabular}{p{0.15\textwidth}p{0.1\textwidth}p{0.13\textwidth}p{0.25\textwidth}p{0.2\textwidth}p{0.1\textwidth}}\toprule
			{\begin{tabular}[c]{@{}l@{}}\textbf{Video}\\ \textbf{}\end{tabular}}    & {\begin{tabular}[c]{@{}l@{}}\textbf{Duration}\\ \textbf{}\end{tabular}}   &{\begin{tabular}[c]{@{}l@{}}\textbf{GT Faces}\\ \textbf{}\end{tabular}}  & {\begin{tabular}[c]{@{}l@{}}\textbf{Insightface}\\ \textbf{(seconds/frame)}\end{tabular}}   & {\begin{tabular}[c]{@{}l@{}}\textbf{Ours}\\ \textbf{(seconds/frame)}\end{tabular}} & {\begin{tabular}[c]{@{}l@{}}\textbf{Speedup}\\ \textbf{factor}\end{tabular}}  \\ \toprule
			\textbf{1}          & 10  & 4  & 9.91      & 2.8    & 3.5x \\
			\textbf{2}          & 10 & 4 & 9.96    & 0.8  & 12.5x  \\
			\textbf{3}          & 60  &4 & 15.8    & 0.3  & 52.7x   \\
			\textbf{4}          & 10 &4      & 10.1      &  0.9  & 11.2x    \\
			\textbf{5}          & 60   &3   & 18.2         & 0.3 & 60.7x
			\\ \cmidrule{4-6}
			& & \textbf{Average}   & 12.8                         & \textbf{1.1}   &   \textbf{28.1x}
			\\\bottomrule
		\end{tabular}
	\end{center}
\end{table}

\section{Conclusion}\label{sec:Con}
The paper presented a method for video face recognition that 
  is shown to be significantly faster and more accurate than the baseline method. 
The method introduced:
  (i) clustering methods to identify image clusters for recognizing faces from different poses,
  (ii) robust tracking with multi-frame processing for occlusions, 
  and (iii) multi-objective optimization to reduce recognition time.

Compared to the baseline method, the final optimized method resulted in 
  speedy recognition times with significant improvements in face recognition accuracy. 
Using face prototype with sampling, the proposed method achieved an accuracy of 
   86.2\% compared to 70.8\% for the baseline system, while running 28.1 times faster 
   than the baseline. In future work, we want to extend our approach to 150 participants
   in about 1,000 hours of videos.
   
\typeout{} 
\newpage
\bibliographystyle{splncs04}
\bibliography{cyprusRef}
\end{document}